\documentclass[acmtog]{acmart}
\acmSubmissionID{xxx}

\usepackage{booktabs} 

\usepackage[ruled]{algorithm2e} 

\SetAlFnt{\small}
\SetAlCapFnt{\small}
\SetAlCapNameFnt{\small}
\SetAlCapHSkip{0pt}

\copyrightyear{2026}
\acmYear{2026}
\setcopyright{rightsretained}
\acmConference[SIGGRAPH '26]{ACM SIGGRAPH 2026 Conference Papers}{July 19--23, 2026}{Los Angeles, CA, USA}
\acmBooktitle{XXX}
\acmDOI{XXX}
\acmISBN{XXX}

\usepackage{graphicx}
\usepackage{amsmath}
\usepackage{mathtools}
\usepackage{caption}
\usepackage{float}
\usepackage{enumitem}
\usepackage{bbm}
\usepackage{wrapfig}
\usepackage{subfigure}
\usepackage{multirow}
\usepackage{comment}
\usepackage{wrapfig}

\usepackage[utf8]{inputenc} 
\usepackage[T1]{fontenc}    
\usepackage{hyperref}       
\usepackage{url}            
\usepackage{booktabs}       
\usepackage{amsfonts}       
\usepackage{nicefrac}       
\usepackage{microtype}      
\usepackage{xcolor}         

\usepackage{cleveref}
\newif\ifdraft
\draftfalse

\definecolor{burntorange}{rgb}{0.8, 0.33, 0.0}
\definecolor{orange}{rgb}{1,0.5,0}
\definecolor{green0}{rgb}{0.1,0.7,0.1}

\DeclareMathOperator*{\argmin}{argmin}

\Crefname{equation}{Eq.}{Eqs.}
\Crefname{figure}{Fig.}{Figs.}
\Crefname{table}{Tab.}{Tabs.}
\Crefname{section}{Sec.}{Secs.}

\newcommand{\parag}[1]{\paragraph{#1}}

\newcommand{\bx}{\mathbf{x}}
\newcommand{\bz}{\mathbf{z}}
\newcommand{\bZ}{\mathbf{Z}}

\newcommand{\mbf}{\mathbf{f}}

\newcommand{\bv}{\mathbf{v}}

\newcommand{\bA}{\mathbf{A}}

\begin{document}
\title{Learning Sewing Patterns via Latent Flow Matching of Implicit Fields}

\author{Cong Cao}
\authornote{Equal contribution.}
\orcid{0009-0001-5989-8367}
\affiliation{%
  \institution{Mohamed bin Zayed University of Artificial Intelligence}
  \country{United Arab Emirates}}
\email{cong.cao@mbzuai.ac.ae}
\author{Ren Li}
\authornotemark[1]
\authornote{Corresponding author is Ren Li (ren.li@mbzuai.ac.ae).}
\orcid{0000-0003-2998-7104}
\affiliation{%
  \institution{Mohamed bin Zayed University of Artificial Intelligence}
  \country{United Arab Emirates}}
\affiliation{%
  \institution{Southern University of Science and Technology}
  \country{China}}
\email{ren.li@mbzuai.ac.ae}
\author{Corentin Dumery}
\orcid{0000-0001-5314-7979}
\affiliation{%
 \institution{École Polytechnique Fédérale de Lausanne}
 \country{Switzerland}}
\email{corentin.dumery@epfl.ch}
\author{Hao Li}
\orcid{0000-0002-4019-3420}
\affiliation{%
  \institution{Pinscreen}
  \country{USA}}
\affiliation{%
  \institution{Mohamed bin Zayed University of Artificial Intelligence}
  \country{United Arab Emirates}}
\email{hao@hao-li.com}


\begin{teaserfigure}
    \includegraphics[width=\textwidth]{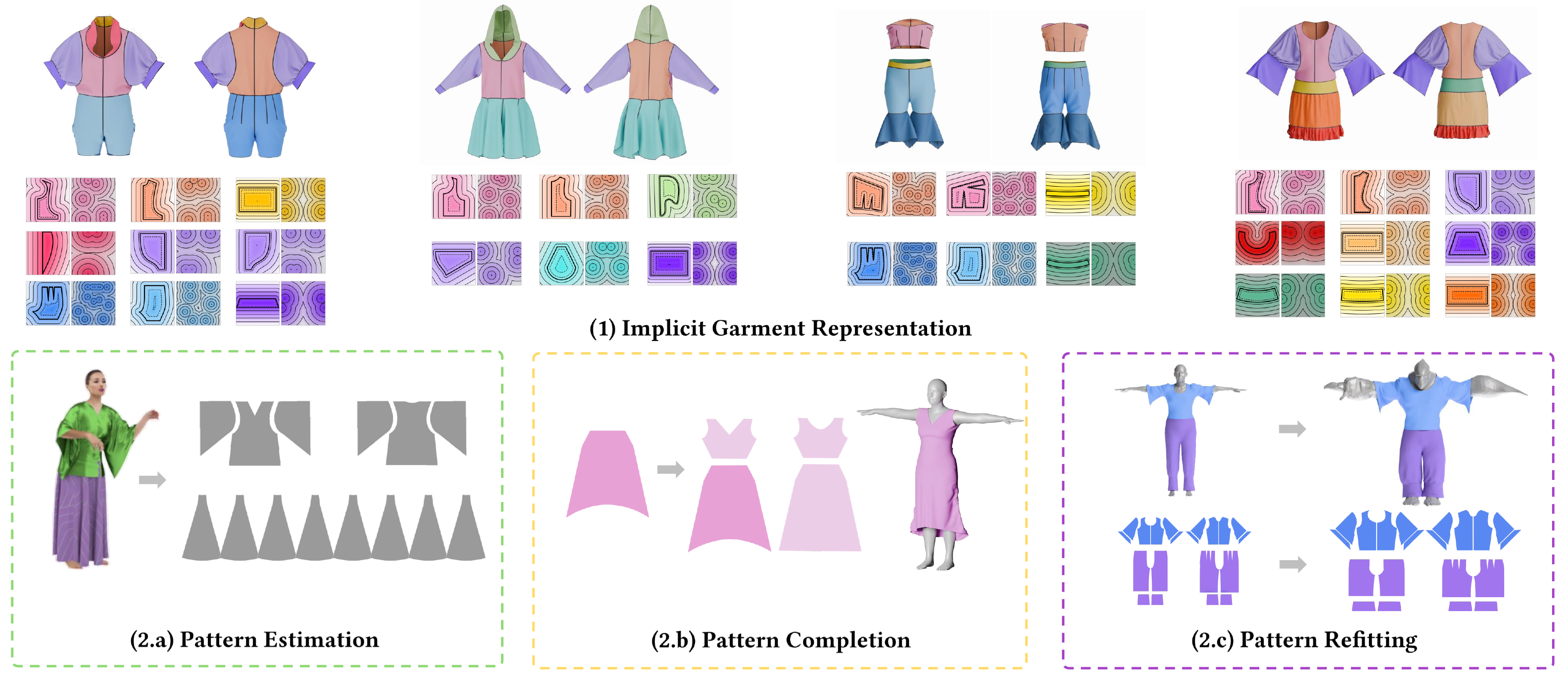}
    \caption{We model and generate sewing patterns using (1) an implicit garment representation, where panels are encoded as continuous distance fields and assembled into garments. Based on this representation, we enable (2.a) sewing pattern estimation from a single image, (2.b) pattern completion from partial panel inputs, and (2.c) pattern refitting for transferring garments across different body shapes.
    }
    \label{fig:teaser}
\end{teaserfigure}


\begin{abstract}

Sewing patterns define the structural foundation of garments and are essential for applications such as fashion design, fabrication, and physical simulation. 
Despite progress in automated pattern generation, accurately modeling sewing patterns remains difficult due to the broad variability in panel geometry and seam arrangements. In this work, we introduce a sewing pattern modeling method based on an implicit representation. We represent each panel using a signed distance field that defines its boundary and an unsigned distance field that identifies seam endpoints, and encode these fields into a continuous latent space that enables differentiable meshing. A latent flow matching model learns distributions over panel combinations in this representation, and a stitching prediction module recovers seam relations from extracted edge segments. This formulation allows accurate modeling and generation of sewing patterns with complex structures. We further show that it can be used to estimate sewing patterns from images with improved accuracy relative to existing approaches, and supports applications such as pattern completion and refitting, providing a practical tool for digital fashion design.
\end{abstract}

\maketitle

\section{Introduction}
\label{sec:intro}
Digital garment modeling is an important component in modern graphics and vision systems, with applications in fashion design, fabrication, realistic physical simulation, telepresence, and immersive VR and AR.  
Sewing patterns serve as a fundamental representation for garment modeling, as they encode the rest shape of garments and specify how garments are assembled in 3D. Operating in a physically meaningful design space, they describe garments as sets of 2D panels defined by their contours and stitching rules. This representation provides direct and interpretable control on garment structural details and stylistic variation.

Despite their wide usage in practice, automatic modeling of sewing patterns remains challenging. Traditional computer-aided design systems support manual drafting and interactive editing, but they require expertise and do not generalize to automated solutions. The difficulty arises from the geometric and topological diversity of sewing patterns. Different garments exhibit panels with various shapes, edge counts and seam configurations. Modeling them requires a method that captures both the geometry of individual panels and the stitching relations that assemble them into full garments.

Recent learning-based approaches attempt to address these challenges. Methods such as~\cite{Yang18f,Wang18i} rely on predefined templates with hand-crafted parameters and use neural networks to estimate them from images. While this simplifies modeling, it restricts representation to fixed template families and shared topology. More recent work adopts vector-based representations that describe sewing patterns as sequences of parameterized lines and curves. \cite{Korosteleva22} predicts curve parameters using a recurrent network, while~\cite{He24b,Nakayama25} quantize vectors into tokens and use transformers to generate patterns. Although these models allow more flexible structures, their autoregressive generation is prone to error accumulation and producing implausible panels. Moreover, vectorized or tokenized representations introduce discontinuities and rely on discrete geometric abstractions that are not differentiable, limiting their use as continuous priors for optimization or reconstruction.
Implicit representations offer a continuous and differentiable alternative. They describe shapes as continuous fields, e.g., signed distance fields (SDF) or occupancy fields, and have shown strong geometric capability. \cite{Li23a} explored the use of implicit fields for sewing pattern modeling, but its reliance on consistent seam labeling and alignment makes adaptation to topology changes difficult. This limits its applicability in practical design scenarios where pattern structures vary widely. 

In this work, we introduce a novel modeling framework for sewing patterns based on an implicit representation to remove these structural restrictions.
Our key idea is to represent each panel using two continuous fields: a signed distance field describing the interior region enclosed by panel edges and an unsigned distance field marking edge endpoints. This design allows us to use a smooth and differentiable function to model panels with arbitrary shapes and edge counts.
A variational autoencoder (VAE) is trained to learn a latent space over these implicit fields for panels, and a flow matching model is then trained in this latent space to learn distributions over panel combinations. 
To assemble panels, stitching relations between panel edges are recovered using a prediction module that evaluates spatial correlations between extracted edge segments. 

This framework not only enables accurate modeling and generation of sewing patterns with complex structures, but also shows better performance in image-based pattern estimation through a conditional generation process. Its generative nature also allows pattern completion by simply using user-provided incomplete patterns as guidance to control the generation trajectory. 
Furthermore, benefiting from the differentiability and continuity of the latent space, our model supports pattern refitting for transferring garments across different body shapes. 
Doing all of these jointly has not been demonstrated in previous works. Codes are at \href{https://github.com/Cao-Cong0/Learning-Sewing-Patterns-via-Latent-Flow-Matching-of-Implicit-Fields}{Github}.

\section{Related Work}
\label{sec:related}

\subsection{Garment Representations}

\parag{Garments as 3D Surfaces}
Early works model garments using mesh templates designed for specific garment types, where variation is introduced via per-vertex deformations or low-dimensional PCA-based shape bases~\cite{Guan12,Patel20,Bhatnagar19}. Garment reconstruction is performed either through optimization with regularization~\cite{Casado22,Zhu22,Liu23b} or learning-based regression~\cite{Danerek17,Jiang20d}. While effective for simple categories, template-based methods rely on manual design and are limited by fixed topology.
To overcome this, recent methods~\cite{Corona21,Li22c,Moon22} adopt implicit surface representations based on SDFs, encoding various garment geometry in a latent space and extracting surfaces via Marching Cubes~\cite{Lorensen87}. Although more flexible, SDF-based modeling of non-watertight garments requires enclosing surfaces in watertight volumes, reducing geometric fidelity and restricting the use of these models in downstream applications. 
Other works~\cite{Guillard22b,DeLuigi23} use unsigned distance functions (UDFs) to avoid this constraint, but accurately learning sharp zero level sets remains challenging and may introduce artifacts and holes in reconstructed surfaces.

\parag{Garments as 2D Sewing Patterns}
Sewing patterns provide an alternative representation that is widely used in garment production and design. A sewing pattern consists of several flat 2D panels together with stitching rules that specify how the panels are assembled. After assembly, real draping or physics-based simulation is then applied to obtain the final 3D garment and its on-body appearance. Despite the assistance of commercial tools such as CLO3D~\cite{Clo3d} and Style3D~\cite{Style3d}, sewing patterns are still designed mainly by hand, which is tedious and requires professional expertise.

Early computational methods~\cite{Yang18f,Wang18i,Liu19k} treat sewing patterns as sets of panel templates defined by a few parameters controlling local dimensions. These parameterized models are simple to use but limited in accuracy and coverage. More recent approaches adopt vector-based pattern representations, where panels are modeled as closed loops of parametric curves such as B-splines or Bézier curves. \cite{Korosteleva21,Korosteleva23} introduce a procedural programming framework that generates sewing patterns at scale and stores curve parameters and stitching logic as JSON configuration files.
Inspired by this progress, methods such as~\cite{Bian25,Zhou25a} use autoregressive models to directly generate JSON configurations compatible with the program execution engine. Other approaches, including~\cite{He24b,Nakayama25,Li25b}, treat sewing patterns as sequences of edge vectors, quantize them into tokens, and use Large Multimodal Models (LMMs) to generate these token sequences autoregressively. Although these strategies enable the synthesis of diverse sewing patterns, their discrete representations and generation process make them infeasible to be used in inverse problems that require gradient-based refinement or optimization in continuous domains.

Another line of work adopts image-based representations for sewing patterns. GarmentImage~\cite{Tatsukawa25} represents sewing patterns as rasterized multi-channel grids that encode panel geometry, topology and placement. Methods like~\cite{Chen22c,Li24a,Li24b,Li25a,Li25} embed both 2D panel shapes and 3D garment geometry into UV positional maps. Different from these explicit representations, ISP~\cite{Li23a} introduces an implicit formulation that models panel shapes and seam boundaries through implicit fields. 
While this approach supports differentiable mesh extraction with respect to the underlying implicit field,
it requires separate models for each pattern topology due to its reliance on consistent boundary labeling and stitching correspondences across garments, which limits its flexibility and scalability.
Our method is also based on the implicit representation, but differs from ISP in that it supports arbitrary numbers of panels, diverse edge structures, and flexible seam configurations within a unified framework. We achieve this by learning a latent space for the implicit fields that jointly model panel geometry and seam endpoints, and by combining it with a latent flow matching strategy and a stitching prediction module to model panel combinations and stitching relationships.

\subsection{Sewing Pattern Recovery}
Existing works recover sewing patterns either from 3D observations such as point clouds or meshes, or from 2D images. For recovery from 3D input, early approaches~\cite{Hasler07,Chen15g,Sharp18,Bang21,Pietroni22} employ template fitting, surface analysis or flattening procedures to infer panel boundaries. More recent methods~\cite{Goto21,Korosteleva22} use learning-based models on point clouds or 3D meshes to predict panel contours and stitching information, leveraging the geometric detail provided by 3D data. 

Estimating sewing patterns from 2D images is particularly challenging because their structural elements are not directly observable. \cite{Jeong15a} predicts garment type and coarse body measurements from a single image and retrieves a sewing pattern from a predefined database. This retrieval-based strategy is inefficient and fails to generalize when the target pose or garment shape differs from the stored examples. Other approaches estimate sewing patterns by optimizing panel parameters through iterative procedures~\cite{Yang18f} or by using encoder-decoder networks~\cite{Wang18i}. These methods require separate templates for different garment types, which limits scalability and restricts the range of garments that can be modeled.
More recent methods employ LMMs~\cite{Liu23i} to regress curve parameters for panel edges~\cite{Liu23j} or to predict discrete edge tokens obtained from pattern quantization~\cite{He24b,Nakayama25}. While these approaches benefit from the expressive power of large models and show promising results, their autoregressive prediction of edges makes them subject to accumulating errors, often producing irregular or undesired panel shapes. In contrast, our method models the entire panel shape through implicit fields and learns panel combinations in a structured latent space, enabling more stable and accurate sewing pattern reconstruction.

\subsection{2D Polygon Representations}

Recent work has explored neural representations for 2D polygonal geometries beyond traditional vertex-edge data structures. Sequence-based approaches~\cite{Carlier20,Reddy21,Liu23} model polygons as ordered vertex coordinates and employ autoregressive architectures for vector graphics generation. To mitigate over-parameterization and resolution limitations, \cite{Polaczek25} adopts implicit neural representations that learn continuous fields to model shape interiors and boundaries. In parallel, graph-based approaches~\cite{Nash20,Yu24} represent polygon vertices as nodes and edges as relational structures, enabling geometric reasoning and topology-aware processing.


\begin{figure*}[ht!]
    \centering
    \includegraphics[width=.99\textwidth]{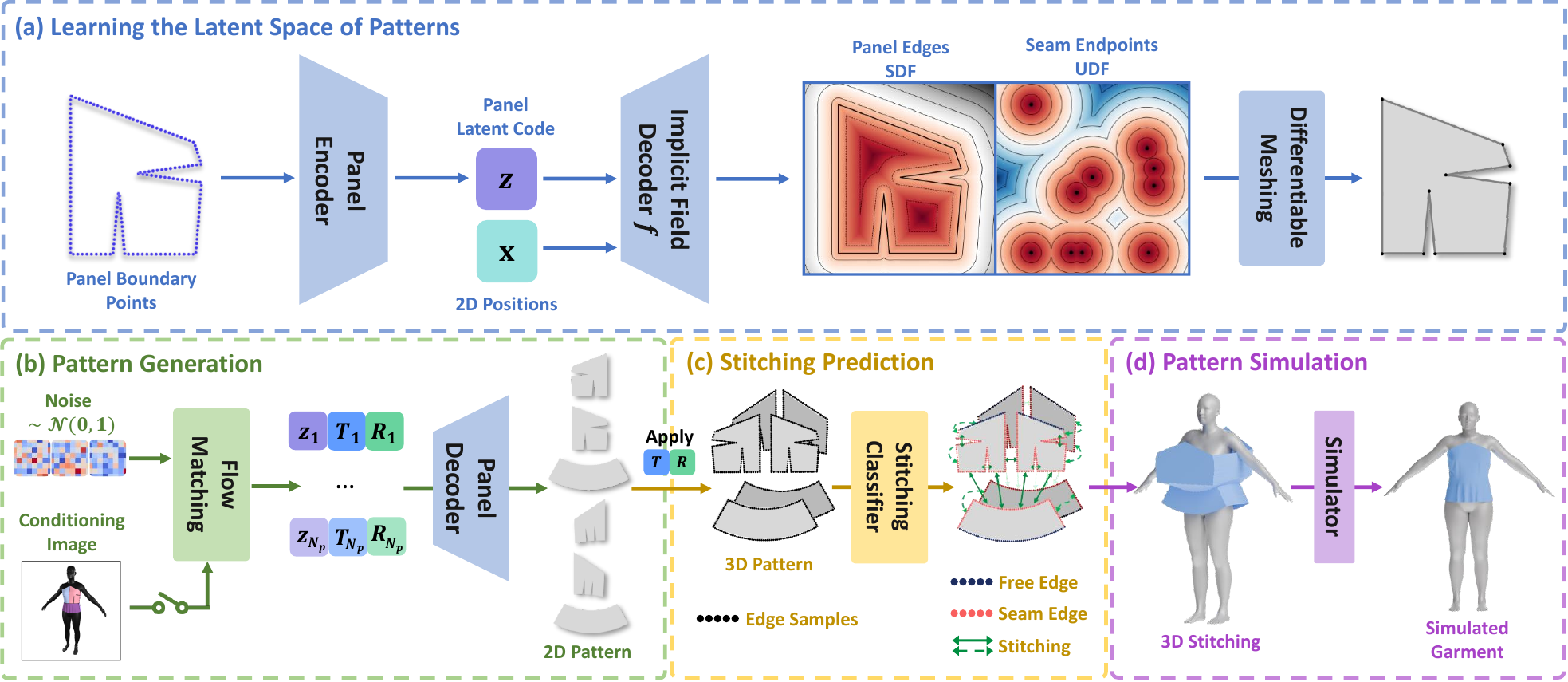}
    \vspace{-0.25cm}
    \caption{\textbf{Overview of the sewing pattern modeling pipeline.}
    (a) We learn a latent space of sewing patterns by encoding panel boundaries into continuous implicit fields, including a signed distance field (SDF) for panel shape and an unsigned distance field (UDF) for edge endpoints. Panel meshes are extracted through differentiable meshing.
    (b) Sewing pattern latent codes $\bz$ are generated by sampling noise and mapping it through a flow matching model, optionally conditioned on an input image. (c) The decoded 2D panels are placed in 3D using predicted translations $T$ and rotations $R$, and stitching relations are recovered by the edge-based classification model. (d) The resulting sewing pattern is then assembled and draped in simulation to produce a 3D garment.}

    \label{fig:pipe}
\end{figure*} 

\section{Method}
\label{sec:method}
A sewing pattern $\mathcal{P} = (P, S)$ is defined as a set of panels $P = \{p_1, ..., p_{N_p}\}$ together with stitching information $S$. Each panel $p_i$ is a closed planar surface in 2D consisting of edges $E = \{e_1, ..., e_{N_e}\}$, where each edge $e_j$ has two endpoints $(o_j, o_j')$. The stitching information $S \subset E \times E$ specifies pairs of edges that should be sewn together during garment assembly.  In practice, sewing patterns exhibit substantial structural variation due to differences in panel layout, edge count and overall topology. In order to accurately capture these attributes, we introduce the modeling framework illustrated in Fig.~\ref{fig:pipe}. It consists of an implicit representation model tailored for individual panels, a latent flow matching model that learns to generate garments in this representation,
and a stitching prediction module that identifies corresponding seams. 
Together, these components form a modeling foundation that supports numerous applications such as pattern generation, estimation and optimization.

\subsection{Modeling Individual Panels}
\label{sec:panel}

\parag{Representation.}
A panel $p$ is formed by $N_e$ consecutive edges $\{e_1, ..., e_{N_e}\}$, where $N_e$ can vary across panels. We use a signed distance field (SDF) to represent the panel shape and an unsigned distance field (UDF) to describe the spatial locations of edge endpoints. Formally, we define an implicit function $f$ as
\begin{equation} \label{eq:implicit}
    (d_c, d_p) = f(\bx, \bz) \; ,
\end{equation}
where $\bx$ is the point in the normalized 2D space $[-1, 1]^2$ and $\bz$ is the latent code that encodes the geometry of the panel. The value $d_c$ is the signed distance to the closed panel boundary, with $d_c < 0$ indicating interior points and $d_c > 0$ indicating exterior points. The value $d_p$ is the unsigned distance to the set of edge endpoints $O = \{o_1, o_1', ..., o_{N_e}, o_{N_e}'\}$, defined as
\begin{equation} \label{eq:udf}
    d_p = \min_{o \in O} \|\bx - o\|_2 \; .
\end{equation}
Thus, the zero level set of the SDF determines the panel contour, while the zero roots of the UDF determine the edge endpoints. These endpoints are later used to segment the panel contour into piecewise edges $\{e_1, ..., e_{N_e}\}$, which is needed for stitching prediction as introduced in Sec. \ref{sec:stitch}.

This representation encodes each panel into a compact latent code and models panels with arbitrary shapes and edge counts through a continuous function $f$. As a result, it preserves differentiability and enables the learned latent space to be used as a prior for optimization tasks.

\parag{Learning the Latent Space.}
The implicit function $f$ of Eq.~\ref{eq:implicit} requires a latent code $\bz$ for each panel. To learn this latent space, we adopt a variational auto-encoding (VAE) strategy. As illustrated in Fig.~\ref{fig:pipe} (a), we use a point transformer~\cite{zhao21b} as the panel encoder and a Multi-layer Perceptron (MLP) as the decoder that implements $f$. For each panel, we randomly sample points along its boundary and use their positions as feature descriptors of the panel geometry. The encoder maps these features to a compact latent code $\bz$, which is subsequently used by the decoder $f$ to predict the SDF and UDF values $(d_c, d_p)$ for any query point $\bx$.
The training objective is
%
\begin{align} \label{eq:vae_loss}
    \mathcal{L} &= \mathcal{L}_{sdf} + \mathcal{L}_{udf} + \lambda_{KL}\,\mathcal{L}_{KL} \; , \\
    \mathcal{L}_{sdf} &= \sum_{\bx} \|d_c(\bx,\bz) - \tilde{d}_c(\bx)\|  + \lambda_{grad}(\|\nabla d_c(\bx,\bz)\|_2 - 1 )^2 \; , \\
    \mathcal{L}_{udf} &= \sum_{\bx} \|d_p(\bx,\bz) - \tilde{d}_p(\bx)\|  + \lambda_{grad}(\|\nabla d_p(\bx,\bz)\|_2 - 1 )^2\; ,
\end{align}
%
where $\tilde{d}_c$ and $\tilde{d}_p$ are the ground truth distance values, $\mathcal{L}_{KL}$ is the KL divergence on the latent code, and $\lambda_{KL}$ and $\lambda_{grad}$ are weighting scalars. 
In contrast to~\cite{Li23a}, which learns latent codes through auto-decoding without an encoder, our use of a panel encoder yields a more structured and geometrically coherent latent space. This improves modeling accuracy and the performance of pattern estimation task, as demonstrated in our experiments.


\begin{wrapfigure}{r}{0.12\textwidth}
    \centering
    \vspace{-0.4cm}
    \includegraphics[width=.11\textwidth]{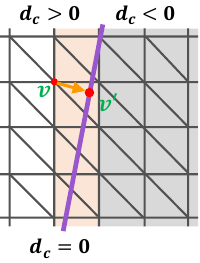}
    \vspace{-0.4cm}
    \label{fig:meshing}
\end{wrapfigure} 

\parag{Inference.}
To extract a triangulated panel mesh and its edge segments from the learned implicit fields, we leverage the gradient of the implicit function $f$. Specifically, we follow the meshing procedure of~\cite{Li23a} to obtain the panel mesh. Given a latent code $\bz$, we initialize a square mesh over the normalized 2D space and evaluate the SDF values $d_c$ at all mesh vertices. As shown in the inline figure, faces (white triangles) whose vertices all have positive $d_c$ are discarded. For faces (orange triangles) that intersect the zero level set (purple line), vertices with positive $d_c$ are projected onto the zero crossing by
\begin{equation}
    v \leftarrow v - d_c(v, \bz)\,\nabla d_c(v, \bz).
\end{equation}
This produces a closed planar mesh $\mathcal{M}$ whose boundary lies precisely on the zero level set of the SDF. The boundary loop $\mathcal{C}$ of $\mathcal{M}$ is then extracted by finding non-paired mesh edges and tracing them.

\begin{figure}[ht!]
    \centering
    \includegraphics[width=.47\textwidth]{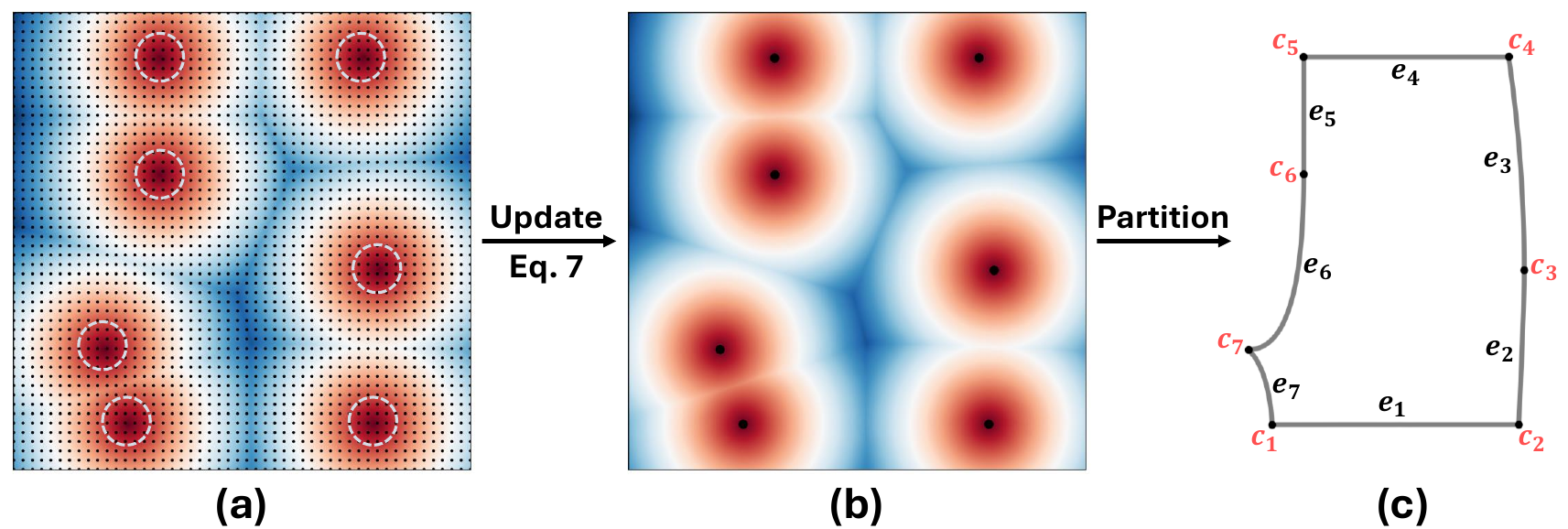}
    \vspace{-0.4cm}
    \caption{\textbf{Inferring edge endpoints.} (a) We evaluate the UDF values $d_p$ on a dense grid (black dots) and retain only points near the zero roots (dark red regions inside dashed circles). (b) These points are iteratively updated using Eq.~\ref{eq:update} to move them toward the zero roots; black dots indicate the updated positions. (c) Cluster centers $c_i$ obtained from the converged points are used as edge endpoints, which partition the panel boundary loop (gray line) into edge segments $e_i$.}
    \label{fig:cluster}
\end{figure}

To segment the boundary loop $\mathcal{C}$ into individual panel edges, we identify edge endpoints using the UDF. Given that the endpoints are discrete and finite, we can define a small margin $\epsilon > 0$ such that the distance between any two unique endpoints $o_i, o_j$ satisfies $\|o_i - o_j\| > \epsilon$. As illustrated in Fig.~\ref{fig:cluster}, we evaluate $d_p$ on a dense grid and retain points with $d_p < \epsilon$.  
For each retained point, we compute an offset $\Delta\bx^*$ that moves it toward a zero root of the UDF, by solving
\begin{equation} \label{eq:update}
    \Delta\bx^* = \argmin_{\Delta\bx} d_p(\bx + \Delta\bx, \bz) \; 
\end{equation}
using gradient descent.
After optimization, we apply DBSCAN~\cite{Ester96} to cluster the converged points $\bx + \Delta\bx^*$ and take the cluster centers as approximated edge endpoints. 
Each detected endpoint is then matched to the closest point on the boundary loop $\mathcal{C}$, which partitions the loop into its constituent edge segments.

Since this meshing process introduces non-differentiable operations, we restore differentiability using the iso-surface gradient formulation from~\cite{Remelli20b,Guillard24a}:
\begin{equation}\label{eq:diff}
    \frac{\partial \bv}{\partial \bz}
    = - \nabla d_c(\bv,\bz)\,
      \frac{\partial d_c}{\partial \bz}(\bv,\bz),
\end{equation}
where $\bv$ denotes the vertices of the extracted panel mesh $\mathcal{M}$.

\subsection{Learning Panel Combinations}
\label{sec:combine}

Different sewing patterns contain different sets and arrangements of panels. To learn the distribution of these combinations, we employ a flow matching model~\cite{Lipman23} built on the implicit panel representation introduced in Sec.~\ref{sec:panel}. Flow matching constructs a continuous trajectory that transports samples from a source distribution $q_0$ to the target distribution $q_1$ by learning a velocity field. 

For each panel in a sewing pattern, we define a feature vector $\tilde{\bz} = [\bz, T, R]$, which concatenates the panel latent code $\bz$, the global translation $T \in \mathbb{R}^3$ and the rotation $R \in \text{SO}(3)$ represented by a quaternion.
As illustrated in Fig.~\ref{fig:pipe} (c), the translation $T$ and rotation $R$ specify how a panel is mapped from its local 2D coordinate frame into the 3D body coordinate system used for draping simulation, following the practice of~\cite{Korosteleva21,Korosteleva23}. A sewing pattern is therefore represented as a set of vectors $\tilde{\bZ} = \{\tilde{\bz}_1, ..., \tilde{\bz}_{N_p}\}$. 
We use a Diffusion Transformer (DiT) architecture~\cite{Peebles23} to learn the velocity field that transports Gaussian noise to the latent distribution of $\tilde{\bZ}$. Each $\tilde{\bz}$ is treated as a token input to the model, and the self-attention layers process these tokens jointly to capture correlations between panels. This allows the network to learn valid panel combinations consistent with real sewing patterns.
To ensure that the learned model is order-agnostic, we remove positional encoding from the input tokens of the DiT.
This property is necessary for tasks like pattern completion, where users usually provide a subset of panels in any order. Without order-agnostic modeling, the generation process would rely on consistent panel ordering, which is not practical in real workflows. 

During training, each pattern is zero-padded to a fixed maximum number of panels.
The flow matching model $U_\theta$ is optimized using
\begin{equation}
    \mathcal{L} =
    \mathbb{E}_{X_0 \sim \mathcal{N},\, t \sim \mathcal{U}}
        \| V_t - U_\theta(X_t, t) \|^2  \; ,
\end{equation}
where $X_0$ is sampled from a standard Gaussian distribution $\mathcal{N}$, $t$ is sampled from the uniform distribution $\mathcal{U}(0,1)$, $X_t$ is the interpolated state between the noise sample $X_0$ and the target $\tilde{\bZ}$, and $V_t$ is the ground truth velocity. 

\subsection{Recovering Stitching}
\label{sec:stitch}

Given a generated pattern $\tilde{\bZ}$ from the latent flow matching model, we infer the corresponding panel meshes using the latent codes and the meshing procedure described in Sec.~\ref{sec:panel}. The inferred meshes are then transformed into the global coordinate frame of a reference body model, i.e. the T-posed SMPL body~\cite{Loper15}, using the predicted translation $T$ and rotation $R$. To recover the stitching relationships required for garment assembly and draping, we first extract edge segments in the global coordinate frame using the edge partitions obtained in Sec.~\ref{sec:panel}. We then use their geometric shapes and spatial positions to predict seam correspondences.

For each edge, we randomly sample $N$ points along it and feed them into a point transformer to obtain an initial edge feature. These features are then processed by a series of self-attention blocks to produce the final edge descriptor $\mbf_e$, which captures both local geometry and global spatial context. Since not all edges participate in stitching, we use a classifier head $\mathcal{H}_{sew}$ to predict whether an edge is a sewing edge based on its descriptor $\mbf_e$.
For a candidate pair of sewing edges $(\mbf_e^i, \mbf_e^j)$, we use the feature disentanglement of~\cite{Lu23} to compute a complementarity logit $\mbf_c^{i,j}$
\begin{equation}
    \mbf_c^{i,j} = \mathcal{G}_{prime}(\mbf_e^i)^{\top} \cdot \bA \cdot \mathcal{G}_{dual}(\mbf_e^j) \; ,
\end{equation}
where $\mathcal{G}_{prime}$ and $\mathcal{G}_{dual}$ are two MLPs that produce primal and dual descriptors to encode the intrinsic local geometry of an edge and its complementary geometry expected from a matching counterpart, respectively. $\bA$ is a learnable square matrix, $\top$ denotes transposition and $\cdot$ denotes matrix multiplication. The logit $\mbf_c^{i,j}$ is finally sent to a sigmoid function $\sigma$ to determine whether a stitch should be assigned to this edge pair.
We train the stitching model using the loss
\begin{equation}
    \mathcal{L} =
    \sum_{i} BE(\mathcal{H}_{sew}(\mbf_e^i),\, l_{sew}^i)
    + \sum_{i}\sum_{j} BE(\sigma(\mbf_c^{i,j}),\, l_{pair}^{i,j}) \; ,
\end{equation}
where $l_{sew}^i$ and $l_{pair}^{i,j}$ are ground truth labels, and $BE$ denotes the binary cross entropy loss.

Since a panel edge in sewing patterns may be stitched to multiple edges, the predicted stitching can contain many-to-many relationships. However, downstream processes such as cloth simulation typically require stitching instructions in a one-to-one form. To satisfy this requirement, we apply a post-processing step similar to the stitch flattening procedure in~\cite{Korosteleva23}. In this step, additional vertices are inserted to subdivide panel edges according to their fractional lengths. The stitch assignments are then propagated to these sub-edges, converting many-to-many or one-to-many matches into valid one-to-one pairs. The edge partitions and stitching relationships are then updated accordingly to reflect the refined structure.


\section{Experiments}

We use the Sewfactory~\cite{Liu23j} and GCD~\cite{Korosteleva24} datasets for pattern learning. Sewing patterns in both datasets are automatically generated by sampling garment design parameters stored in JSON template files. They cover a wide range of garment variety, e.g. shirts, pants, jumpsuits, skirts and dresses. We parse each template to extract closed panel contours and edge endpoints, normalize them into a unified 2D space, and use them to generate SDF and UDF training samples consisting of 2D query points and their distance values.
Both datasets provide sewing patterns paired with images of simulated garments draped on human bodies. GCD images are direct renderings of the 3D simulations, while SewFactory images are further processed using a human texture synthesis pipeline to produce photorealistic appearances. Since Sewfactory does not provide an official train-test split, we adopt a custom split for training and evaluation. For GCD, we follow the data splits defined in AIpparel~\cite{Nakayama25}. Additional training details are provided in the supplementary materials. 

Our method provides a complete representation of sewing patterns. As illustrated in Fig.~\ref{fig:teaser} (1), it enables diverse pattern and garment generation by sampling and mapping the random noise to the latent space through the denoising process.
Beyond this basic capability, we further evaluate the method on three downstream applications to demonstrate its effectiveness in practical scenarios.


\begin{table*}[ht]
  \caption{\textbf{Quantitative evaluation of panel quality.} $\uparrow$: the higher the better; $\downarrow$: the lower the better.} \label{tab:panel}
    \begin{center}
    \vspace{-0.35cm}
    \scalebox{0.8}{
     \begin{tabular}{c | c | c | c | c | c | c }
        \toprule
          Sewfactory & IoU $\uparrow$ & Trans L2 $\downarrow$ & Rot L2 $\downarrow$ & \#Panel $\uparrow$ & \#Edge $\uparrow$ & DSR $\uparrow$ \\
         \midrule
         Sewformer & 0.821 & 0.367 & \textbf{0.008} & 86.06\% & 97.35\%  & 63.1\%\\
         Ours & \textbf{0.847}  & \textbf{0.296} & 0.015 & \textbf{88.92\%} & \textbf{98.63\%} & \textbf{81.0\%} \\
        \bottomrule
    \end{tabular}
    }
    ~~~~~
    \scalebox{0.8}{
     \begin{tabular}{c | c | c | c | c | c | c}
        \toprule
          GCD & IoU $\uparrow$ & Trans L2 $\downarrow$ & Rot L2 $\downarrow$ & \#Panel $\uparrow$ & \#Edge $\uparrow$ & DSR $\uparrow$\\
         \midrule
         AIpparel & 0.834 & 1.783 & \textbf{0.004} & 91.39\% & 86.53\% & 62.6\% \\
         Ours & \textbf{0.892}  & \textbf{1.274} & 0.0167 & \textbf{91.99}\% & \textbf{93.26\%} & \textbf{69.0\%} \\
        \bottomrule
    \end{tabular}
    }
      \end{center}
      
\end{table*}

\begin{table}[ht]
    \begin{center}
    \caption{\textbf{Quantitative evaluation of stitching prediction.} $\uparrow$: the higher the better.} \label{tab:stitch_tab}
    \vspace{-0.35cm}
    \scalebox{0.7}{
     \begin{tabular}{c | c | c | c }
        \toprule
          Sewfactory & F1 $\uparrow$ & Precision $\uparrow$ & Recall $\uparrow$  \\
         \midrule
         Sewformer & 0.956 & 0.967 & 0.951  \\
         Ours & \textbf{0.993}  & \textbf{0.996} & \textbf{0.991}  \\
        \bottomrule
    \end{tabular}
    }
    ~
    \scalebox{0.7}{
     \begin{tabular}{c | c | c | c }
        \toprule
          GCD & F1 $\uparrow$ & Precision $\uparrow$ & Recall $\uparrow$  \\
         \midrule
         AIpparel & 0.821 & 0.822 & 0.821  \\
         Ours & \textbf{0.981}  & \textbf{0.982} & \textbf{0.981}  \\
        \bottomrule
    \end{tabular}
    }
      \end{center}
      
\end{table}

\subsection{Estimating Sewing Patterns from Images}\label{sec:pattern_est}

To enable image-based sewing pattern estimation, we extend the flow matching model $U_\theta$ introduced in Sec.~\ref{sec:combine} to an image-conditioned generation setting.
Inspired by~\cite{Wu24a}, we condition the latent generation process on both pixel-level and semantic-level image features to better align the panel latent space with visual evidence from the input image. We use the DINO-v3~\cite{Simeoni25} to extract pixel-level image features $\mbf_I^p$, represented as a collection of patch tokens, and the CLIP visual encoder~\cite{Radford21} to extract a semantic embedding $\mbf_I^s$. The pixel-level feature $\mbf_I^p$ is concatenated with the noisy latent tokens $X_t$ and processed through a self-attention layer, allowing the model to capture their intrinsic relationships and refine the noisy latent representation. The semantic embedding $\mbf_I^s$ is then injected into the refined noise latent through a cross-attention layer, enabling the integration of high-level contextual cues from the image into the latent generation process.
The conditional flow matching is trained by minimizing 
\begin{equation}
    \mathcal{L} =
    \mathbb{E}_{X_0 \sim \mathcal{N},\, t \sim \mathcal{U}}
        \| V_t - U_\theta(X_t, t, \mbf_I^p, \mbf_I^s) \|^2 \; ,
\end{equation}
where the model now predicts a velocity field conditioned on image features.

We train this model using paired image and sewing pattern data from Sewfactory and GCD. We compare our approach with two state-of-the-art methods, Sewformer~\cite{Liu23j} and AIpparel~\cite{Nakayama25}, and provide the ablation studies in supplementary materials. Sewformer uses a transformer model to regress parameters for each panel edge, while AIpparel finetunes a large vision-language model to predict discretized edge tokens in an autoregressive manner. Since Sewformer does not release the train-test split used for its pretrained model, we retrain it using our data splits. For AIpparel, we use the released pretrained model.

\parag{Evaluation Metrics}
To quantitatively evaluate sewing pattern estimation, we use metrics similar to those in prior works~\cite{Korosteleva22,Liu23j,Nakayama25,Li25a}. 
We compute:  
1) \textbf{Panel IoU}: the Intersection-over-Union (IoU) between the ground truth and predicted panels modeled as 2D polygons, measuring panel shape accuracy;  
2) \textbf{Rot L2} and \textbf{Trans L2}: the L2 distances between predicted and ground truth rotations and translations, respectively, with \textbf{Trans L2} measured in cm;  
3) \textbf{\#Panel}: the accuracy of the predicted number of panels within each pattern;  
4) \textbf{\#Edge}: the accuracy of the number of edges within each correctly predicted panel;  
5) \textbf{Precision, Recall and F1 score} of stitching prediction, evaluating the correctness of the recovered stitching relationships;
6) \textbf{Draping success rate (DSR)}: the percentage of generated sewing patterns that can be successfully assembled and draped in simulation.

\subsubsection{Evaluation on Panel Quality}

Table~\ref{tab:panel} reports quantitative results on panel quality for sewing patterns predicted from single view images. Our method achieves better performance than Sewformer and AIpparel in terms of Panel IoU, Trans L2, \#Panel, \#Edge and DSR. In particular, the improvements in Panel IoU and \#Edge indicate that our estimations more accurately capture panel shapes. We attribute this to our implicit panel representation, which models the complete panel shape using a single latent code. In contrast, Sewformer and AIpparel represent sewing patterns as sequences of edges and generate them sequentially, which makes their predictions prone to error accumulation. We observe slightly higher rotation error (Rot L2) compared to the baselines. However, since rotation is mainly used to place panels in 3D space for subsequent draping simulation which is not sensitive to small deviations, the increased Rot L2 does not lead to noticeable degradation in practice. 

\subsubsection{Evaluation on Stitching Accuracy}
Table~\ref{tab:stitch_tab} presents the evaluation results for stitching accuracy. Our method outperforms both baselines by a large margin across all stitching metrics. AIpparel predicts stitching relations autoregressively and only conditions on previously generated edges, while Sewformer relies mainly on pairwise edge feature similarity. In contrast, our method explicitly leverages both local geometric properties and global spatial context of edges when predicting stitching relations, which leads to improved accuracy.

\subsubsection{Qualitative Results}
\begin{figure*}[ht!]
    \centering
    \includegraphics[width=1.\textwidth]{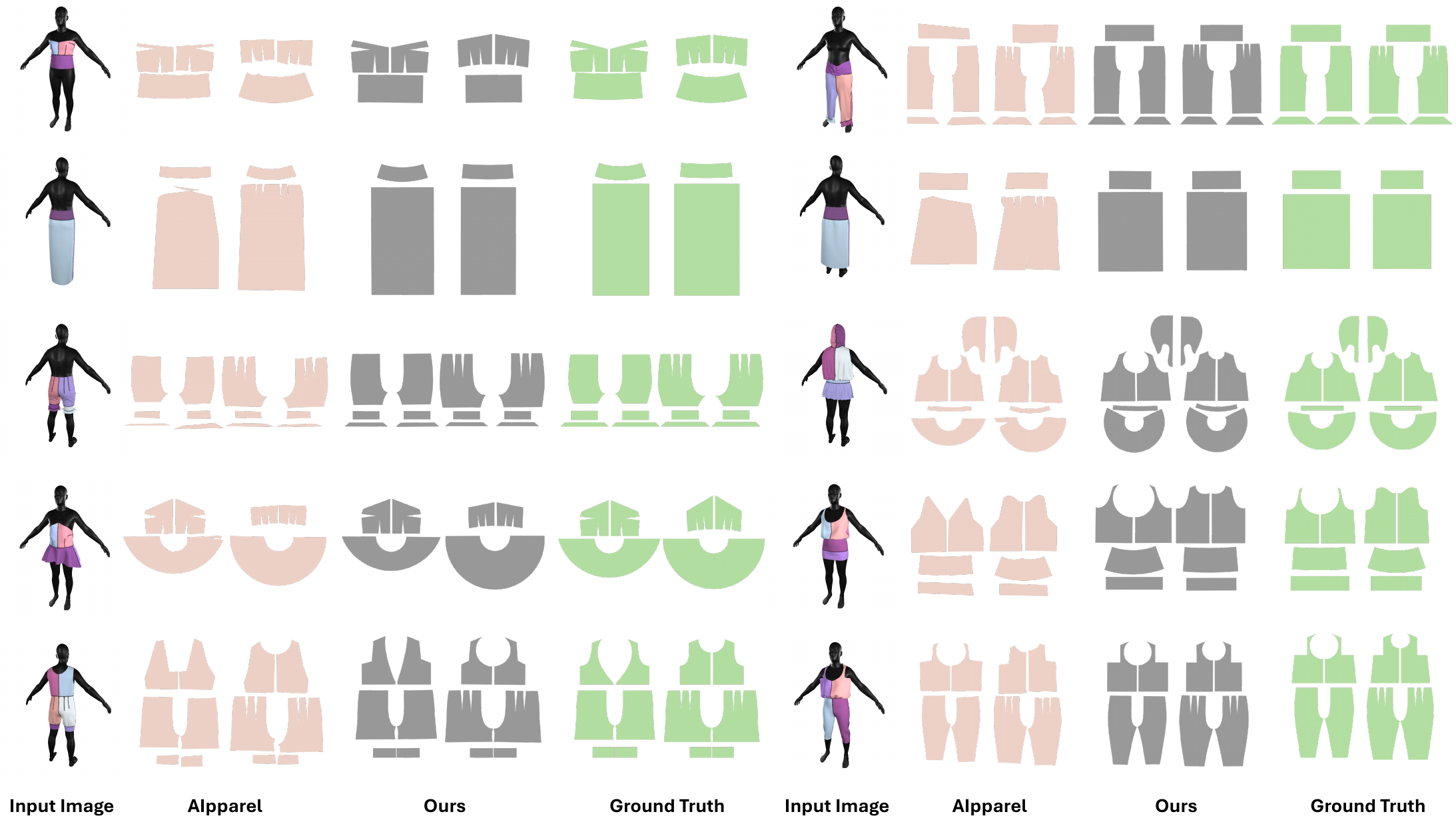}
    \caption{\textbf{Qualitative comparison on the GCD dataset.} From left to right: input image, results from AIpparel, results from our method, and ground-truth sewing patterns. AIpparel often produces extra edges and distorted panel shapes due to its autoregressive edge generation.
    In contrast, our method produces results with more accurate panel geometry and structure, closely matching the ground truth across diverse garment types.}
    \label{fig:compare_GCD}
\end{figure*}

Fig.~\ref{fig:compare_sew} and Fig.~\ref{fig:compare_GCD} present qualitative comparisons between our method and the baseline approaches. The baseline methods often produce inaccurate sewing patterns with distorted panel shapes and invalid additional panels. In contrast, our method reconstructs sewing patterns whose panel geometry and overall structure more closely align with the ground truth.

\subsection{Pattern Refitting}\label{refit}
Since our method is based on an implicit panel representation that supports differentiable mesh extraction (Eq.~\ref{eq:diff}), we demonstrate that it can be used as a prior to refit sewing patterns for transferring generated garments across different body shapes and sizes.

Given a source garment $\mathcal{M}_{s}$ fitted to a source body $\mathcal{B}_{s}$ together with its corresponding sewing pattern, 
the goal of pattern refitting is to optimize the sewing pattern such that the resulting garment is adapted to the shape of the target body $\mathcal{B}_{t}$.
To this end, we first generate a target refitted garment mesh $\mathcal{M}_{t}$ for the target body $\mathcal{B}_{t}$ following the procedure of~\cite{deGoes20,Chen25}. This produces a target garment in 3D space that reflects the desired fit on $\mathcal{B}_{t}$, but has no associated 2D patterns and thus cannot directly be fabricated or used in pattern-based design workflows.

Our differentiable garment representation enables us to leverage a differentiable neural simulator~\cite{Grigorev24} to optimize the sewing pattern so that the simulated garment $\mathcal{M}$ matches the target mesh $\mathcal{M}_{t}$ as closely as possible. The optimization minimizes
\begin{equation}\label{eq:refit}
\mathcal{L} =
\lambda_{c} \mathcal{L}_{CD}(\mathcal{M}, \mathcal{M}_{t})
+ \lambda_{l} \mathcal{L}_{\mathrm{lap}}(\mathcal{M})
+ \lambda_{s} \mathcal{L}_{\mathrm{seam}}(\mathcal{M}) \; 
\end{equation}
where $\lambda_{c}$, $\lambda_{l}$, and $\lambda_{s}$ are scalar weights. The term $\mathcal{L}_{CD}$ measures the discrepancy between the simulated garment and the target garment, $\mathcal{L}_{\mathrm{lap}}$ enforces surface smoothness through Laplacian regularization, and $\mathcal{L}_{\mathrm{seam}}$ encourages consistency between corresponding seam lengths.
As the simulated garment $\mathcal{M}$ is a function of the sewing pattern, which is parameterized by the latent codes $\bz$, we minimize the above objective with respect to $\bz$ using the chain rule. More details can be found in supplementary materials. 

\begin{figure*}[th!]
    \centering
    \includegraphics[width=.99\textwidth]{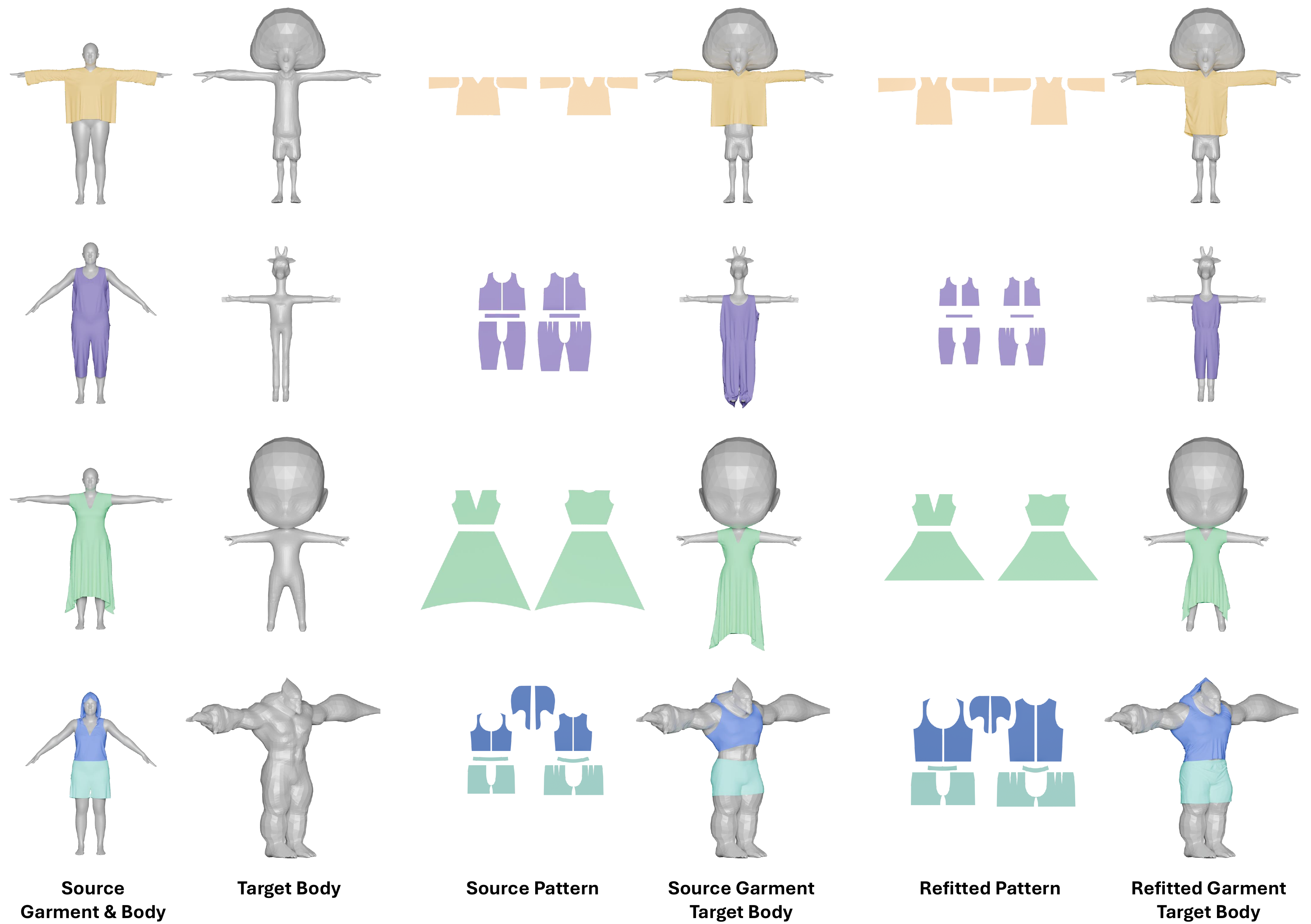}
    \caption{\textbf{Pattern refitting across different body shapes.} Given a source garment $\mathcal{M}_s$ and its sewing pattern defined on a source body $\mathcal{B}_s$, our method refits the pattern to adapt it to a target body shape $\mathcal{B}_t$. The refitted sewing patterns produce garments that conform to the target body while preserving the structural characteristics of the source design.}
    \label{fig:refit}
\end{figure*}

Optimizing in the learned latent space constrains the refitting process to produce valid panel shapes, avoiding distortions that commonly arise when directly optimizing edge control points.
As shown in Fig.~\ref{fig:teaser} and Fig.~\ref{fig:refit}, our method can refit sewing patterns to bodies with different proportions and sizes. The refitted patterns produce garments that conform well to the target body shape while preserving the original design structure.
To quantitatively evaluate it, we conduct an experiment in which panel shapes are optimized directly through edge control points using Eq.~\ref{eq:refit}, and compare this with our optimization in the latent space. We use triangle quality as a measure of geometric validity following \cite{Chen25}. The triangle quality scores are 0.781 for edge control optimization and 0.875 for latent-space optimization, demonstrating the improved geometric consistency of our approach.

\subsection{Pattern Completion}

Given partial panel specifications provided by users, such as a set of $m$ panels represented by contours or binary mask images that indicate desired panel shapes, our framework is able to complete them into a full sewing pattern.
We first recover their corresponding latent codes by enforcing zero SDF values on the panel boundary $\mathcal{C}$:
\begin{equation} \label{eq:recover_z}
    \bz^* = \argmin_{\bz}
    \sum_{\bx \in \mathcal{C}} d_c(\bx, \bz)^2
    + \lambda_{\bz} \|\bz\|_2 \; ,
\end{equation}
where $\lambda_{\bz}$ is a regularization weight.
The recovered latent codes $\{\bz_i^*\}_{i=1}^m$ are then used to guide the generation trajectory of the flow-matching model $U_\theta$ following the control strategy of~\cite{Patel25}. Specifically, given a noisy state $X_t$ at time step $t$, we refine the trajectory by updating $X_t$ through gradient descent to solve
\begin{equation} \label{eq:guidance}
    X_t^* = \argmin_{X_t}
    \sum_{i=1}^m \| z_{i, \hat{X}_1} - \bz_i^* \|_2 \; ,
\end{equation}
where $\hat{X}_1 = X_t + (1-t)\cdot U_\theta(X_t, t)$ denotes the estimated denoised sample, and $z_{i, \hat{X}_1}$ is the latent code corresponding to the $i$-th panel in $\hat{X}_1$.
This guidance encourages the generated result to match $\{\bz_i^*\}_{i=1}^m$, while leveraging the learned distribution prior of $U_\theta$ to fill in the remaining panels in a coherent and plausible manner. More
details can be found in supplementary materials.

As illustrated in Fig.~\ref{fig:GCD_completion} (a), our method can recover plausible complete sewing patterns when provided with different numbers of panels. In addition, 
we can also recover their translations and rotations for panel layout using the same strategy, and infer stitching relationships using the stitching classifier. This enables assembling and draping the garment in simulation, as shown in Fig.~\ref{fig:GCD_completion} (b).
\section{Conclusion}

We introduced an implicit modeling framework for automatic sewing pattern modeling. The method represents panels using continuous distance fields and combines a learned latent space with flow matching and stitching prediction to enable flexible modeling of panel combinations and stitching relations. The framework supports both pattern generation and image-based sewing pattern estimation, and further enables downstream applications such as pattern completion and refitting across body shapes. Experimental results demonstrate improved accuracy and robustness compared to existing approaches.

\parag{Limitations and Future Work}

While our method is capable of modeling diverse sewing patterns, it has several limitations. 
First, although the proposed representation is differentiable with respect to panel geometry, the pattern topology remains discrete. Thus, the current model does not support optimization across sewing patterns with different panel counts or topologies. Enabling differentiable pattern topology is an important direction for future work.

Second, our method focuses on modeling outer panel boundaries, but does not explicitly handle interior seam edges or internal components such as pockets or decorative elements. Supporting such structures would require extending the representation to encode interior constraints and their relationships to panel geometry.

Third, as a data-driven generative model, generation beyond the training distribution remains a limitation. During sampling, the model may hallucinate out-of-distribution latent codes, resulting in irregular distance fields and invalid panels. In addition, the stitching classifier is not perfect; prediction errors may propagate to later stages, leading to incorrect seam assignments and edge partitions.

\begin{figure*}[ht]
	\centering
		\begin{minipage}{.485\textwidth}
        \centering
        \includegraphics[width=.99\textwidth]{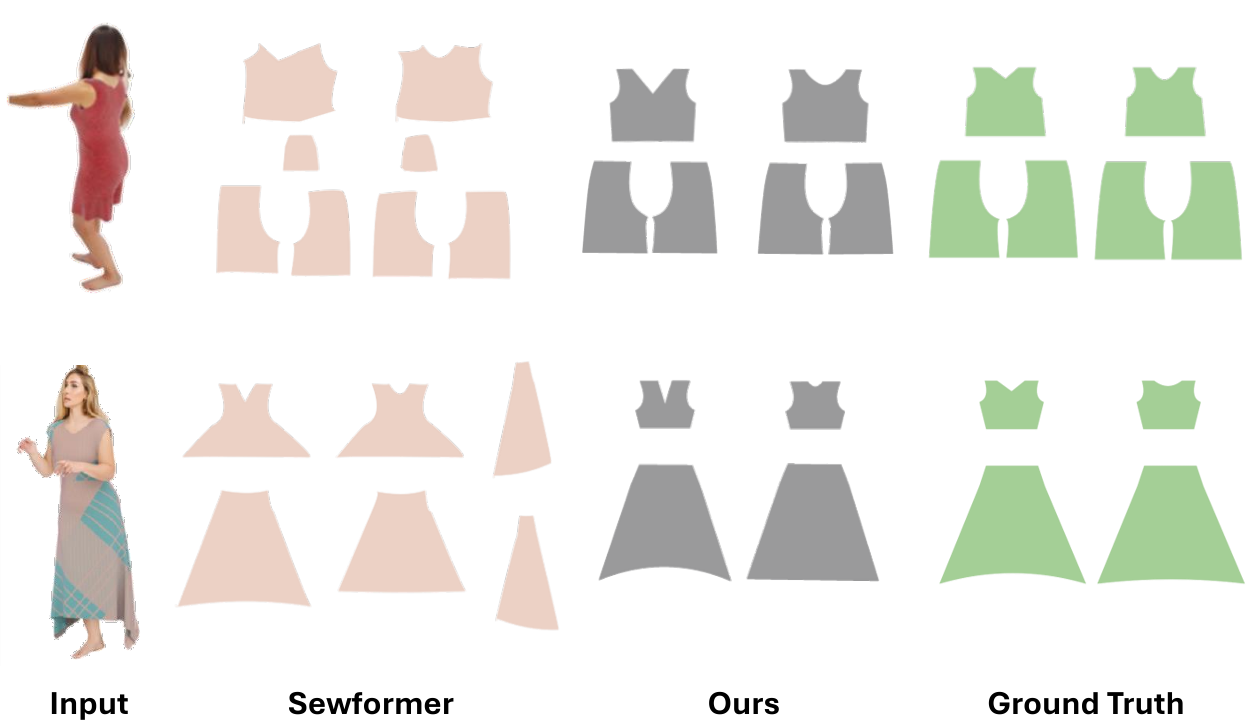}
        \caption{\textbf{Qualitative comparison on the SewFactory dataset.} From left to right: input image, results from Sewformer, results from our method, and ground-truth sewing patterns. Compared to the Sewformer, our method reconstructs panel shapes and overall pattern structure that more closely match the ground truth, while avoiding distorted panels and invalid components.}
    \label{fig:compare_sew}
    \end{minipage}
	 \hspace{2mm}
	\begin{minipage}{.485\textwidth}
		\centering
        \includegraphics[width=.99\textwidth]{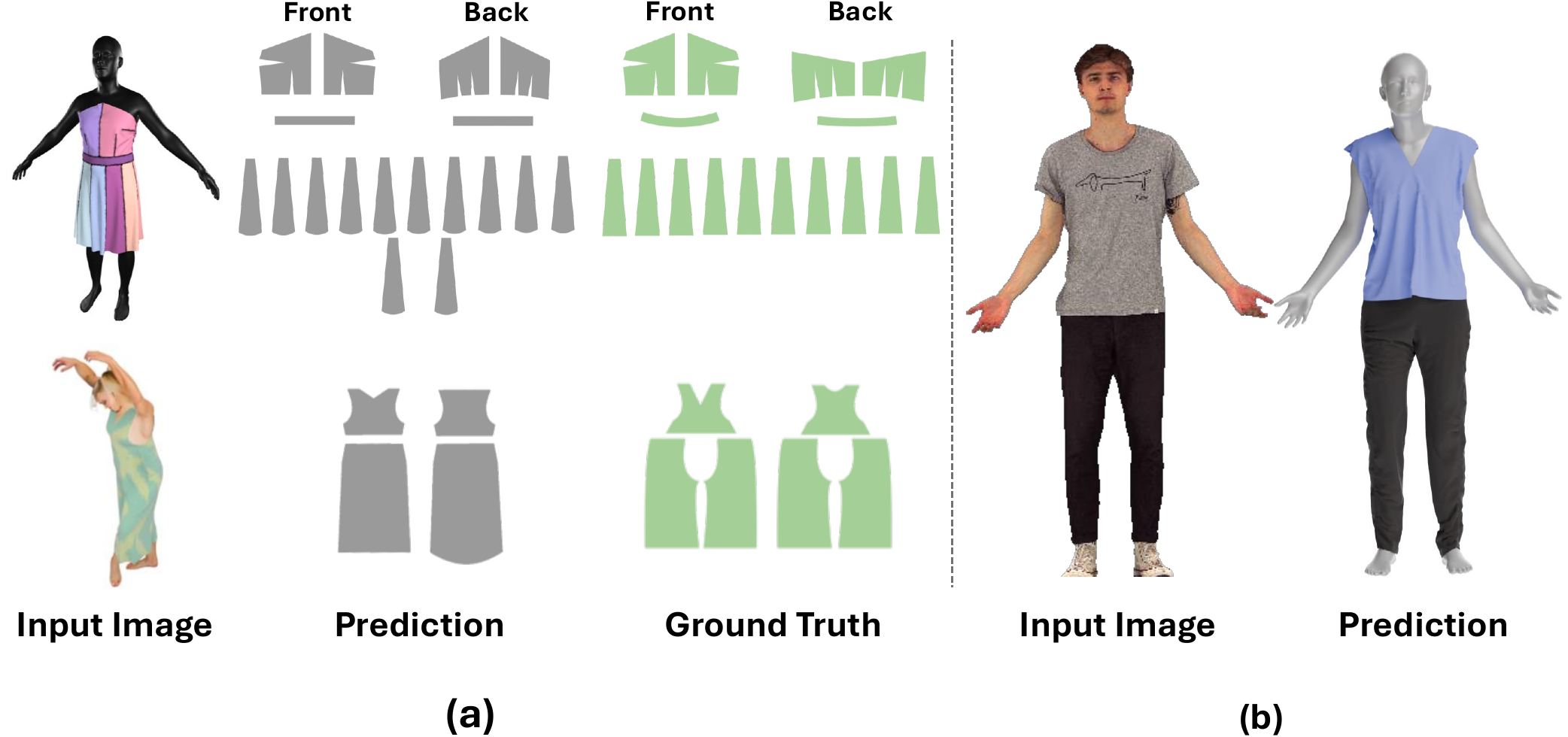}
        \caption{\textbf{Failure cases.} (a) Examples where sewing pattern estimation from single-view images is inaccurate. Occlusions and ambiguous garment designs can lead to incorrect panel structures or extra panels. (b) Due to the synthetic-to-real domain gap, our model trained on synthetic data may produce incorrect estimations when applied to real images (examples from~\cite{Wang24}).}
        \label{fig:limitation}
	\end{minipage}
\end{figure*}

Finally, sewing pattern estimation from single-view images is inherently ambiguous due to limited visual information and design variability. Occlusions and viewpoint constraints can lead to incorrect panel counts or structures (Fig.~\ref{fig:limitation} (a)).
Moreover, as trained primarily on synthetic datasets, the model may degrade on real images due to domain gaps in appearance, lighting, and clothing (Fig.~\ref{fig:limitation} (b)). 
Addressing this issue would require either collecting large-scale real-world datasets with ground-truth sewing patterns or developing more robust domain adaptation strategies.

\begin{acks}
This work is supported by the Metaverse Center Grant from theMBZUAI Research Office.
\end{acks}

\bibliographystyle{ACM-Reference-Format}
\bibliography{string,geom,graphics,learning,vision,biomed,misc,new_ref}

\clearpage

\begin{figure*}[ht!]
    \centering
    \includegraphics[width=.84\textwidth]{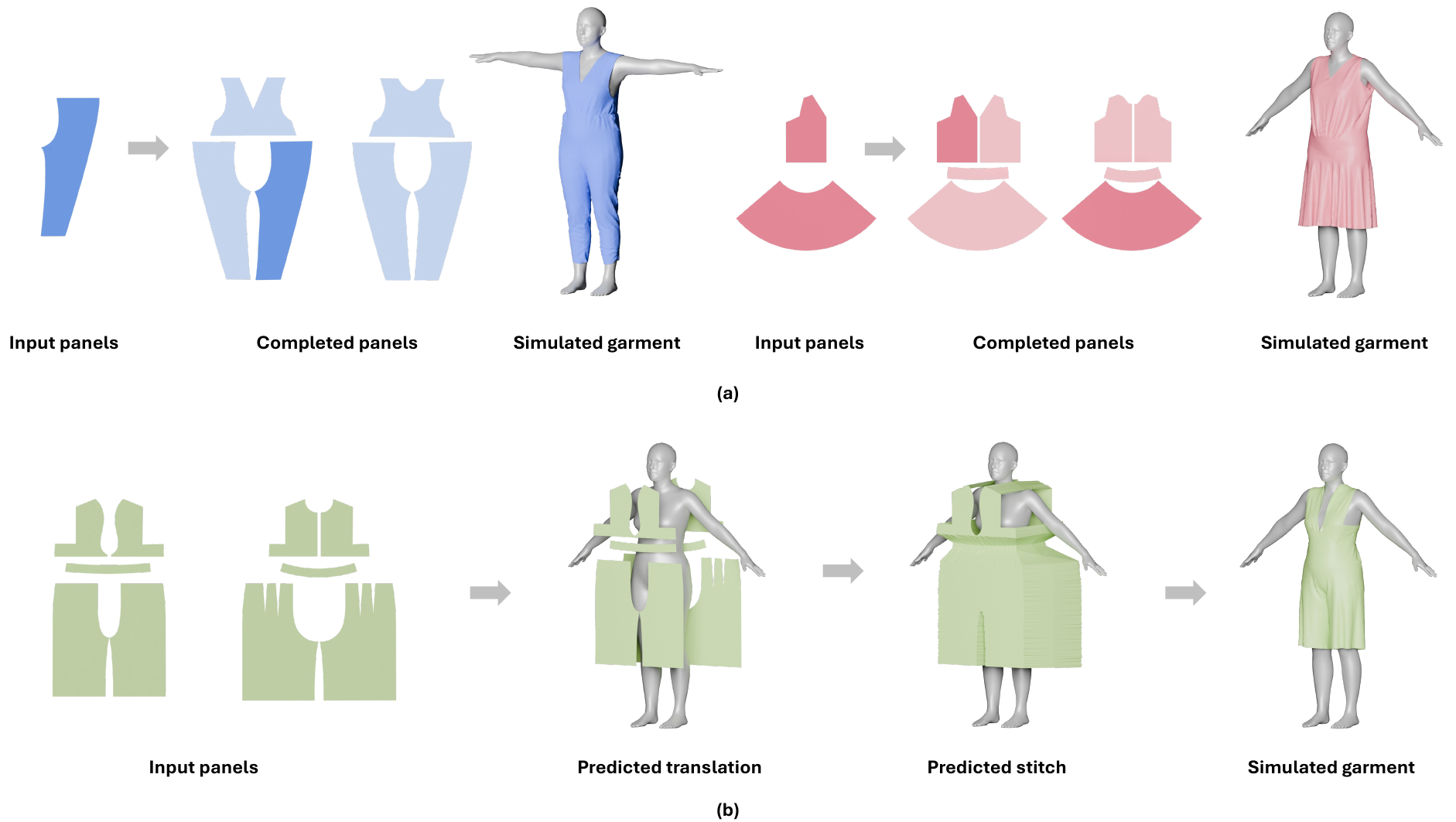}
    \caption{\textbf{Pattern completion.} (a) Given partial panel inputs, our method completes the missing panels to form a full sewing pattern. (b) The model also recovers panel layout and stitching relationships, allowing the completed patterns to be assembled and draped in simulation to produce 3D garments. 
    }
    \label{fig:GCD_completion}
\end{figure*} 
\clearpage

\end{document}